\documentclass[conference]{IEEEtran}
\usepackage{hyperref}
\usepackage{caption}
\usepackage[export]{adjustbox}
\usepackage{tabularx}
\usepackage{booktabs}
\usepackage{xcolor}
\usepackage{algorithm}
\usepackage{subfig}
\usepackage[binary-units]{siunitx}
\usepackage[noend]{algpseudocode}
\def\tablename{Table}

\begin{document}
\title{
Automated Design Space Exploration for optimised Deployment of DNN on Arm Cortex-A CPUs
}

 \author{\IEEEauthorblockN{Miguel de Prado}
 \IEEEauthorblockA{miguel.deprado@he-arch.ch}
 \and
 \IEEEauthorblockN{Andrew Mundy
 \IEEEauthorrefmark{3}}
 \IEEEauthorblockA{andrew.mundy@arm.com}
 \and
 \IEEEauthorblockN{Rabia Saeed
 \IEEEauthorrefmark{1}}
 \IEEEauthorblockA{Rabia.Saeed@he-arc.ch}
 \and
  \IEEEauthorblockN{Maurizo Denna
 \IEEEauthorrefmark{4}}
 \IEEEauthorblockA{maurizio.denna@nviso.ch}
 \and
 \IEEEauthorblockN{\hspace{5cm}Nuria Pazos
 \IEEEauthorrefmark{1}}
 \IEEEauthorblockA{\hspace{5cm}Nuria.Pazos@he-arc.ch}
 \and
 \IEEEauthorblockN{\hspace{-11cm}Luca Benini
 \IEEEauthorrefmark{2}}
 \IEEEauthorblockA{\hspace{-11cm}lbenini@iis.ee.ethz.ch}
 \and
 \IEEEauthorblockA{\hspace{3cm}\IEEEauthorrefmark{1} He-Arc Ingenierie, HES-SO
 \IEEEauthorrefmark{2}Integrated System Lab, ETH Zurich
 \IEEEauthorrefmark{3}Arm Ltd.
 \IEEEauthorrefmark{4}Nviso}
 \vspace{-1cm}
 }

\maketitle

\begin{abstract}
The spread of deep learning on embedded devices has prompted the development of numerous methods to optimise the deployment of deep neural networks (DNN). Works have mainly focused on: \textit{i)} efficient DNN architectures, \textit{ii)} network optimisation techniques such as pruning and quantisation, \textit{iii)} optimised algorithms to speed up the execution of the most computational intensive layers and, \textit{iv)} dedicated hardware to accelerate the data flow and computation. However, there is a lack of research on cross-level optimisation as the space of approaches becomes too large to test and obtain a globally optimised solution. Thus, leading to suboptimal deployment in terms of latency, accuracy, and memory. 

In this work, we first detail and analyse the methods to improve the deployment of DNNs across the different levels of software optimisation. Building on this knowledge, we present an automated exploration framework to ease the deployment of DNNs. The framework relies on a Reinforcement Learning search that, combined with a deep learning inference framework, automatically explores the design space and learns an optimised solution that speeds up the performance and reduces the memory on embedded CPU platforms. Thus, we present a set of results for state-of-the-art DNNs on a range of Arm Cortex-A CPU platforms achieving up to $4\times$ improvement in performance and over $2\times$ reduction in memory with negligible loss in accuracy with respect to the BLAS floating-point implementation.

\end{abstract}
 
%
\IEEEpeerreviewmaketitle

\vspace{0.6cm}
\section{\textbf{Introduction}}
Deep learning has grown quickly in the last years, achieving remarkable results in computer vision~\cite{rowley1998neural, howard2017mobilenets, yamashita2018convolutional} and speech recognition~\cite{graves2013speech, kepuska2017comparing}. 
Recently, the focus has shifted towards the deployment of such DNNs on resource-constrained devices to make the myriad of devices on the edge intelligent. In contrast to cloud environments, edge devices are often severely constrained in terms of computing power, memory, and energy consumption, which is available to a given application. These constraints hamper deployment of deep learning solutions to edge devices and require innovation in the design of deep learning systems (or neural network architectures), and in the software which executes them.

\begin{figure}[t!]
   \centering
   \includegraphics[width=0.95\linewidth, right]{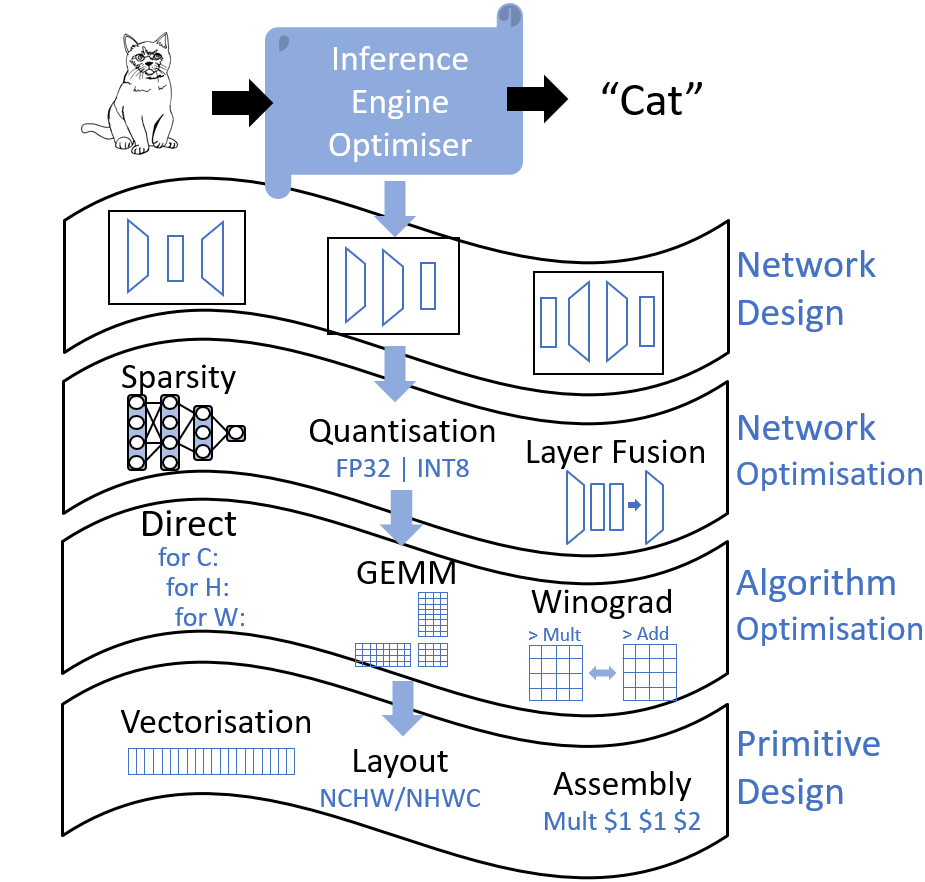}
   \caption{\small\textbf{Optimisation categories} for the deployment of DNNs at different levels of the software stack.}
   \label{opts}
\end{figure}

Numerous research works have focused on optimizing the deployment of DNN through the development of \textit{i)} efficient DNN architectures such as MobileNets~\cite{howard2017mobilenets}, SqueezeNet~\cite{iandola2016SqueezeNet}, including hardware-aware neural architecture search (NAS)~\cite{yang2018netadapt, wu2019fbnet}, \textit{ii)} optimisation techniques such as pruning and quantisation~\cite{wang2019deep, he2018amc, han2015deep, stock2019and}, \textit{iii)} optimised algorithms to speedup the execution of the most computational layers, e.g., general matrix multiplication (GEMM)~\cite{anderson2017low} or Winograd~\cite{maji2019efficient} and, \textit{iv)} dedicated hardware to accelerate the data flow and parallelise the computation~\cite{oh2004gpu, zhang2015optimizing, andri2016yodann}. 

There is, however, a lack of research on cross-level optimisation and design space exploration (DSE) for a complete end-to-end solution~\cite{ai-pipeline}. The space of approaches for DNN deployment becomes too large to fully explore and obtain an optimal implementation as each layer of the neural network may be executed following a different optimisation technique, algorithm, or even in a different processor, resulting in a different performance, memory footprint or power consumption. The complexity of exploring and combining the wide variety of design options usually results in a sub-optimal deployment solution~\cite{anderson2018optimal}.

\vspace{0.4cm}
The objective of this work is to ease the deployment of pre-trained DNNs for industrial applications by automatically exploring the design space and finding an optimised solution to speed up the performance and reduce the memory on embedded CPU platforms. To that end, we employ LPDNN~\cite{de2018quenn}, a deep learning framework that enables the deployment and inference of DNN implementations. We focus on software optimisation for the deployment on Arm CPU cores as these represent the majority of processors on mobile and IoT devices and have extensive support for DNN inference~\cite{wu2019machine}. Our work is complementary to DNN architecture design for further deployment optimisation and could also be applied to dedicated hardware. Our contributions are the following: 

\begin{itemize}
    \item We analyse methods to improve the deployment of DNNs across different levels of software optimisation (Section~\ref{background}) and introduce the range of techniques provided by LPDNN to optimise DNN inference (Section~\ref{engine}).
    \item We present QS-DNN (Section~\ref{qs-dnn}), an automatic exploration framework based on Reinforcement Learning (RL), that, combined with LPDNN, finds an optimised combination of design options that speeds up DNN inference and reduces memory for a target platform.
    \item We present a set of results for state-of-the-art DNNs on a wide range of Arm Cortex-A CPU platforms that cover the current spectrum of the deployment on mobile devices (Section~\ref{results}). Finally, we show a comparison between RL and several other search algorithms when evaluated on our deployment design space.
\end{itemize}

Next, we present the deployment optimisation of DNNs across different levels of the software stack.

\section{\textbf{Background: Deployment Optimisation of Deep Neural Networks}} \label{background}
Given the constraints imposed on edge devices, namely, relatively limited compute performance, small memory capacities, and thermal and power consumption restrictions, there are several goals for which one may choose to optimise. For example, one might decide to sacrifice neural network inference latency to reduce overall power consumption, or to stay within a more limited memory capacity. Depending on the goal of the optimisation, neural networks present a range of software optimisation opportunities. We divide these opportunities into several broad categories as shown in Fig.~\ref{opts}:

\subsection{\textbf{Network Design}}
We define network design optimisation to be the set of techniques that tailor the structure of a network before, or during training, to improve the latency or cost of network inference. Examples of this are MobileNet-V1/V2~\cite{howard2017mobilenets,sandler2018MobileNetv2}, SqueezeNet~\cite{iandola2016SqueezeNet} and, ShuffleNet~\cite{zhang2018shufflenet} which were manually shaped thanks to the expertise of the authors. A set of newer works introduced the neural architecture search (NAS) as a technique to reduce the high-level human knowledge needed for the conception of such architectures. Examples of this are MNASnet~\cite{tan2019mnasnet}, FbNet~\cite{wu2019fbnet}, and Lemonade~\cite{elsken2018efficient} which used hardware-aware NAS via reinforcement learning, evolutionary algorithms or gradient-based methods to discover neural network structures with both good accuracy and high performance. 

Distillation is another technique where a neural network teacher can transfer its learned knowledge to student networks. Students are constructed to present lower computational complexity and memory cost than their teachers. The student imitates the teacher over a training dataset and obtains high accuracy while reducing the complexity of the neural network. Works implementing this technique are~\cite{hinton2015distilling, frosst2017distilling, crowley2018moonshine}.

\subsection{\textbf{Network Optimisation}}
In contrast to neural network design, network optimisation techniques take an existing network and modify the way it is represented, for example by exploiting lower-precision data types (quantisation), the inherent sparsity of weights and activations (pruning) or fusion of linear operations (layer fusion). Techniques in this category may require a neural network to be retrained to deal with a loss of accuracy, or to enforce a particular structure upon sparse data. However, significant reductions in memory consumption can be achieved.

\subsubsection{\textbf{Pruning}}
Explicit representation of sparsity can result in less computational work being wasted on needless computations (such as multiplication by zero). There are two categories of sparsity: unstructured and structured. The former can achieve a higher degree of sparsity over the tensor but involves a high overhead to decode the sparse format at inference time, making this approach more suitable for specialised accelerators~\cite{han2015deep, elafrou2018sparsex, nurvitadhi2017can}. The latter exploits the inherent structure of the data. One example of this is forcing a reduction of feature maps within a layer resulting in a dense but smaller tensor~\cite{fedorov2019sparse, crowley2018pruning, anwar2017structured}.

\subsubsection{\textbf{Quantisation}}
Quantisation is a compression method that reduces the storage cost of a variable by employing reduced-numerical precision. This improves the arithmetic intensity of neural network inference by increasing the amount of computational work which can be performed for a given amount of memory traffic. Some of the first works addressing hardware-oriented quantisation of DNNs were Ristretto~\cite{gysel2016ristretto} and~\cite{lin2016fixed}, which analysed the effect of quantising weights, biases, and activations for each layer. DNN quantisation gained much attention and evolved quickly towards: \textit{i)} reducing the bitwidth down to binary networks~\cite{rastegari2016xnor, choi2018pact} and, \textit{ii)} techniques to find a suitable trade-off between compression and accuracy where autoML methods represent the SoA for mixed-precision inference~\cite{wang2019haq, dong2019hawq, wu2018mixed}. We refer the reader to extensive analyses for efficient quantisation deployment, which are provided by~\cite{krishnamoorthi2018quantizing, wang2019deep}. 

All the mentioned works provide quantisation methods or tools that involve training or fine-tuning the DNNs to push the limits of quantisation as well as a large training dataset. There are, however, several works that provide tools for post-training (direct) quantisation achieving 8-bit~\cite{zhao2019improving, nagel2019data, quantnvidia} or even 4-bit~\cite{banner2019post, nagel2020up} inference with minimal loss in accuracy, making them very attractive for any user to deploy DNNs efficiently on embedded devices. 


\subsubsection{\textbf{Layer fusion}}
Layer fusion can improve the memory traffic of a DNN as several linear operations in a neural network graph can be fused into a single one -- avoiding repeatedly writing and rereading the same area of a tensor. In general terms, the fusion of two consecutive layers approximately halves the memory traffic associated with the combination. Examples of this are merging the batch normalisation and scale layer or the activation and concatenation layer into the previous convolution or fully connected layer. Further memory optimisations can be achieved by different layers sharing the same memory space if there is no dependency between them, e.g., in-place computation or network-memory pool.

\subsection{\textbf{Algorithm Optimisation}}
Once a network has been designed (and possibly optimised through application of quantisation, sparsity, or fusion), execution of the network can be optimised through modification of the way in which layers of the network are implemented. We mainly focus on the performance optimization of convolutions since these comprise the lion’s share of the computation work contained in a neural network. There are several ways in which convolution can be performed: direct convolution; a number of the several approaches that exploit a General Matrix-matrix Multiplication (GEMM) call; or by one of many “fast-convolution” methods like Winograd convolution. Each of these methods has its own trade-offs.

\subsubsection{\textbf{Direct}}
A naive approach to implementing convolution on CPU is to directly implement the six-nested for loop which describes convolution.  
Although direct convolutions incur no memory overhead, its usage is rare since it is difficult to express the algorithm in a way that extracts much performance from CPU architectures~\cite{dukhan2019indirect}. Implementations can be improved by keeping the some of the weights, inputs, or outputs resident in registers -- especially for a small number of parameters~\cite{directconv} -- and reordering the layout and loops to optimise data reuse~\cite{zhang2018high}.

\subsubsection{\textbf{GEMM-based}}
Use of GEMM is attractive to accelerate convolutions since there exist a wide range of fast implementations provided by highly optimised BLAS libraries such as OpenBLAS~\cite{openblas} or BLIS~\cite{van2015blis} capable of exploiting the SIMD instructions of the Armv8-A architecture.
The prototypical approach to constructing a GEMM-backed convolution is to use the \texttt{im2col} or \texttt{im2row} algorithms to construct a ``patch'' matrix, which can be multiplied by a matrix representing the convolution weights to form the final output matrix.
It should be noted, however, that while the amount of work performed by the GEMM is equivalent to direct convolution, the memory footprint is $k^{2}$ larger (where $k$ is the size of the kernel). This significant memory overhead has led to research into more memory efficient GEMM-backed convolutions. Examples of which are the \texttt{kn2row} technique~\cite{anderson2017low}, and indirect GEMM~\cite{dukhan2019indirect} approach -- neither of these techniques reduce the arithmetic cost of performing a convolution, although they do avoid the cost of rearranging the data into \texttt{im2col} or \texttt{im2row} form.

\subsubsection{\textbf{Winograd}}
Winograd Convolution~\cite{lavin2016fast} can help to address the problem of the high arithmetic cost of convolution. These algorithms help to reduce the overall compute complexity of convolution by transforming the convolution into another domain where the number of required “strong” operations (such as multiplication) is reduced at the expense of an increase in the number of “weak” operations (such as addition). Implementations of these algorithms are well suited to low power embedded systems, as the resources and power budget are very limited. By contrast, they have a higher cost in memory consumption and accuracy~\cite{maji2019efficient}.

\subsection{\textbf{Primitive design}}
Finally, the lowest level of software instantiating a neural network can be optimised to make better use of the hardware upon which it is executed. All of the algorithms described in the previous sections feature at least one loop, which will be executed many thousands of times during neural network inference. Ensuring that this innermost loop is implemented as well as possible is vital to achieving good overall performance. Optimisations at this level of abstraction can vary from changing the layout of data in memory, through changing how vectorised instructions are used to process the layer, to writing assembly implementations of the kernels to extract maximum performance from specific processors. 

\subsubsection{\textbf{Data layout}}
Ensuring that operands are laid out to achieve proper use of the processor cache hierarchy, and easy exploitation of vectorized execution may produce significant improvements in performance~\cite{maji2019efficient}. Indeed, several works, including~\cite{anderson2018optimal, de2019learning}, apply several algorithms to find an optimised selection of data layout for each layer of a DNNs.

\subsubsection{\textbf{Vectorisation}}
It is crucial that the vector (Single Instruction Multiple Data -- SIMD) instructions provided by the Instruction Set Architecture (ISA) are used to make the most of processor throughput. Examples of works leveraging vectorisation for the optimisation of convolutions on GPU or CPU are~\cite{rovder2019optimising, van2015blis, maji2019efficient}. 

\subsubsection{\textbf{Assembly code}}
Modern compilers, while good at ensuring general-purpose code can be compiled into fairly efficient assembly, have some drawbacks. Writing assembly code by hand can allow the programmer to perform optimisations missed by the compiler and allows for a much greater degree of control of the final binary.

\subsection{\textbf{Discussion}}
The vast majority of the works presented above focused on specific optimisations for DNNs without taking into account the trade-offs at different levels of the software stack. We draw inspiration from Anderson et al.~\cite{anderson2018optimal} who use PBQP to optimize inference time by selecting suitable backends. However, they only profile the latency for convolutional layers without addressing other optimisations at network level, e.g., quantisation. In this work, we provide a broader picture and show the various steps of the optimisation for the deployment of DNNs on CPU (Section \ref{background}\textit{-B, -C and -D}). Furthermore, we present an automatic exploration framework, based on Reinforcement Learning, that searches through different design option and analyses several DNNs on a range of embedded platforms while trading off metrics like latency, memory, or accuracy. Thereby, we can find a solution that, for instance, can answer the following questions: What DNN shall I use? What are the best optimisation techniques that I can follow? How can I obtain a fast implementation under a certain memory or accuracy constraints?  

\section{\textbf{Deep learning inference framework}} \label{engine}
We form part of a European collaboration to bring deep learning methods to any party who would like to take up deep learning solutions in an industrial environment~\cite{llewellynn2017bonseyes}. In this context, a deep learning framework (LPDNN) has been developed~\cite{de2018quenn} to produce efficient and tunable code that enables and maximizes the portability among platforms. In this work, we introduce the range of techniques provided by LPDNN to optimise the deployment of DNNs. Besides, we address the integration of the LPDNN into our search environment to tightly couple empirical measurements of a heterogeneous platform to a learning-based search.

\subsection{\textbf{Architecture}}
One of the main goals of LPDNN is the portability and flexibility of AI applications across platforms. LPDNN's core comprises a set of CPU dependency-free functions which can be complemented by specific-platform acceleration libraries, such as Arm Compute Library~\cite{ArmCL} (ArmCL) and cuDNN~\cite{chetlur2014cudnn}, to generate an optimised implementation for the system. LPDNN contains a modular and hierarchical architecture that supports multiple libraries and optimisations at the same time. This flexibility allows us to experiment with several algorithms for a particular layer or blocks and execute them with the most suitable implementation according to the network architecture, target platform, and desired accuracy and performance specification.

\subsection{\textbf{Optimisations}} \label{lpdnn-opt}
We follow the structure given in Section~\ref{background}, and show the software optimisations that LPDNN contains at various levels:

\subsubsection{\textbf{Network optimisation}} \label{memory-opt}
LPDNN provides efficient inference with integer arithmetic as it supports post-training quantisation for both weights and activations. Weights can be directly quantised to 8-bit integer while the activations require a validation set to determine their dynamic range. LPDNN uses this information in a step during the compilation process (code generation of the DNN) where it calculates the scale and offset for both symmetric and asymmetric quantisation methods. The scale value can be further tuned to reduce the loss of information by minimising the KL divergence between the quantised and original distribution as introduced in~\cite{quantnvidia}. LPDNN supports both per layer and per channel quantisation. However, due to the lack of support by the acceleration libraries for channel quantisation, we only focus and show results for the former.

Several other optimisations are performed in LPDNN:
\begin{itemize}
    \item \textbf{Static layer fusion:} Fusion of linear operations to reduce the neural network graph at build time. LPDNN supports the fusion of the Bnorm and scale layers into the previous convolution or fully connected layer. 
    \item \textbf{Runtime layer fusion:} The execution of two or more layers is performed in a single pass with a significant reduction in memory traffic. LPDNN supports the fusion of the activation and concatenation layers into the previous convolution.
    \item \textbf{In-place computation:} Layers such as activation, reshape or, flatten may store the output result directly on the memory allocated for the input, which halves the memory allocations for a layer.
    \item \textbf{Memory pool:} Layers whose execution does not overlap and who do not have data dependencies, share the same memory, which -- due to the sequential nature of most DNNs -- notably reduces overall memory footprint.
\end{itemize}

\subsubsection{\textbf{Algorithm optimisation}}
LPDNN integrates a set of acceleration libraries including, OpenBLAS, BLIS, NNPACK and, ArmCL, that provide optimised algorithms for the execution of DNNs on CPUs. LPDNN leverages the algorithms provided by the libraries and may execute each layer of the network with a different algorithm. Further, LPDNN also uses a lower-level interface of the Arm Compute Library that we refer to as LPDNN-Arm library. It supports both floating-point 32-bit (FP32) and integer 8-bit (INT8) operations and provides special optimisations for the following layers:
\begin{itemize}
    \item \textbf{Standard convolution:} LPDNN integrates a FP32 fast convolution implementations that relies on Winograd for 3x3, 5x5 and linear kernels (originally from~\cite{maji2019efficient}), and a vectorised GEMM implementation for all kernels, including the common 1x1, for both FP32 and INT8.
    \item \textbf{Depthwise Convolution:} Despite containing relatevely little computational work, is challenging to implement efficiently due to its memory-bound nature. LPDNN's exploits all the reuse presented by the algorithm by carefully mapping to both the SIMD instructions and the cache for both FP32 and INT8.
    \item \textbf{Others:} LPDNN also optimises pooling, element-wise and, activation layers by providing vectorisation for both FP32 and INT8 implementations.
\end{itemize}

\subsubsection{\textbf{Primitive design}}
In Section~\ref{background}, we noted three different elements of primitive design which can be combined to build optimised kernels to implement neural network algorithms.
These were \emph{data layout}, \emph{vectorisation}, and 
\emph{assembly code}.

The first two of these are neatly tied together: the order in which data is stored suggests the vectorisation approach taken and vice versa.
For example, when implementing a vectorised convolution one may decide to operate on several channels of data simultaneously, in which case storing channels contiguously facilitates easier use of the data.
%
We have determined empirically that, on CPUs, it is often better (both easier and more performant) to write kernels which operate on multiple channels simultaneously~\cite[\S~2.1]{maji2019efficient}.
Consequently, the majority of optimised kernels in LPDNN operate of NHWC-ordered data (where N stands for the number of \emph{batches}, H for the \emph{height} of the tensor, W for the \emph{width} and C for the number of \emph{channels}).
However, it can still be beneficial to support data in other formats (as shown in Section~\ref{results}) -- for example, when the channel count is low it is better to make use of data-reuse across the plane of a convolution.

We noted above that it can often be worth hand-writing assembly code implementations for key kernels, rather than relying on the compiler.
There are a few reasons for this, largely stemming from wanting finer-grain control over register allocation, instruction selection and scheduling than is possible from use of compiler intrinsics.
LPDNN integrates several hand-optimised vendor kernels covering algorithms such as GEMM and depthwise convolution.

\section{Learning-based Search Engine}
\label{sec:learning-based-search-engine}

In this section, we address the design space for the deployment's optimisation of DNNs (Section \ref{background}) and we propose Reinforcement Learning as a solution.

\begin{figure}[t!]
   \centering
   \includegraphics[width=0.9\linewidth]{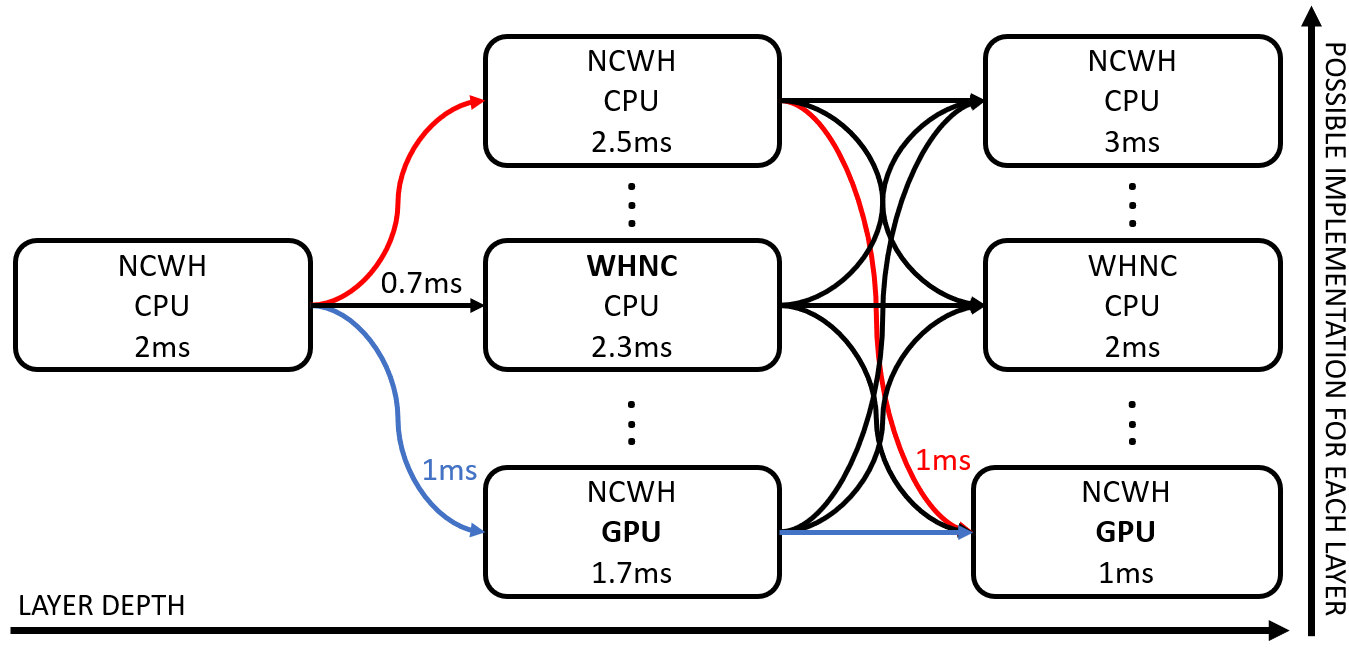}
   \caption{\small\textbf{Conversion penalties.} 3-layer example network showing a local minimum due to incompatibility penalties between implementations (arrow time). Red path solution represents a local minimum as it contains the fastest intermediate implementation while the the blue path is the fastest overall.}
   \label{layers}
   \vspace{-0.4cm}
\end{figure}

\subsection{\textbf{Problem formulation}}
We address the problem of finding an optimised DNN deployment strategy to speed up the performance and reduce the memory by selecting among the myriad of implementations to execute each network layer. The problem is not as trivial as to select the fastest implementation for each layer. Each layer may follow a different optimisation technique or be executed by various acceleration libraries, which, in turn, might provide several algorithms and implementations. Each of these may have a different layout, data type, or even be executed on another processor, which might not correspond to those from the previous and following layers. Therefore, incompatibilities between layers arise needing data copy/transformation layers, e.g., conversion from FP32 to INT8 or from NHWC to NCHW layout, that incur in extra penalties.

The complexity of exploring and combining the wide variety of design options becomes too large to test exhaustively, which usually results in a sub-optimal solution~\cite{anderson2018optimal} due to local minimum selection (see Fig.~\ref{layers}). The number of combinations within a network, which is the design space to explore, grows exponentially with the depth (D), i.e, number of layers, having as base the branching factor (B), i.e., different implementations for each layer. Hence, the design space size for a network would be 
\begin{math} 
    \mathcal{O(B^{D})}
\end{math} as the worst case. If we take SqueezeNet as an example, a relatively small network, and assume we have seven different implementations for each layer, that would give us a design space of $7^{66} \approx \num{6e55}$.

This is a non trivial problem and therefore, a careful search must be carried out to select the right set of deployment options that, combined and assuming the conversion penalties, yields the most suitable solution for a given goal, e.g., latency, accuracy or memory. We can consider our design space as a weighted graph where each implementation is a vertex ($\mathcal{V}$) and the connections between consecutive layers are the edges ($\mathcal{E}$). We define the cost (weight) of moving from one vertex to another as the implementation cost of the next vertex plus the transformation penalty between both implementations (if any), see Fig. \ref{layers}.

Thus, we evaluate several search algorithms for our deployment design space and show the results in Section \ref{results}. We include \textit{Random} and \textit{Direct} searches, transversal searches such as \textit{Dijkstra} and \textit{A*}, which includes heuristics in the search, and Reinforcement Learning (RL). We select RL as the search algorithm for our solution, which we describe as follows.

\subsection{\textbf{Reinforcement Learning Approach}}
RL lends itself perfectly to exploring large design spaces due to its sample-based approach and far-sighted accumulative reward~\cite{li2017deep, sutton1998introduction}. Consider the network space exploration as a Markov Decision Process (MDP) containing an agent.
We are interested in learning a function that optimises the agent's behavior or policy $\pi(a_t|s_t)$, i.e., mapping from state $s_t$ to actions $a_t$, without modeling the environment and only relying on the reward function. Q-learning~\cite{watkins1989learning} fits well this description as it is a model-free and value-based implementation, having the policy implicit in the value function. The action-value function $q_\pi$ is the expected return $G_t$ in a state $s_t$ taking an action $a_t$:
\begin{equation}
	q_\pi\left(s,a\right) = E_\pi\left[G_t | s_t = s, a_t = a\right]
\end{equation}
The objective of Q-learning is to maximize the total reward:
\begin{math}
R_T = \sum_{t=0}^{\infty}\gamma^{t}r_{t}
\end{math}
where $r_t$ is an individual reward and $\gamma$ is the discounted factor for successive states. Besides, Q-learning is an off-policy implementation, that is, it may follow a behavior policy $\mathcal{A}$ while targeting a greedy policy $\mathcal{B}$. Following Bellman's equation, we can iteratively update the action-value function ($Q=q_\pi$) as follows:
\begin{equation}
Q(s_t,a_t)=Q_{s_t, a_t}(1-\alpha) + \alpha\left[r_t + \gamma\max_{a}Q(s_{t+1},a)\right]
\label{bellman}
\end{equation}

\begin{table}[t!]
\begin{center}
\begin{tabular}{ll}
\toprule
\multicolumn{1}{c}{State Parameters} & \multicolumn{1}{c}{Definition} \\
\midrule
Layer type & Any layer, e.g., convolution, pooling  \\
Layer depth & Position of the layer in the network\\
Acceleration library & Name of the library \\
Algorithm & Routine type \\
Algorithm config & Sub-routine or lowering method \\
Data type & Any type, e.g., FP32, FP16, INT8 \\
Data layout & Any layout, e.g., NCHW, NHWC \\
Target hardware core & CPU, GPU, FPGA. \\
\bottomrule
\end{tabular}
\end{center}
\caption{\small\textbf{State Space.} Parameters define the execution implementation of a layer on a target platform.}
\label{state_space}
\vspace{-0.4cm}
\end{table}

\subsection{\textbf{Search Engine}}
We consider an agent whose aim is to learn the optimal path among a large but finite set of states $\mathcal{S}$ i.e., layer representations, employing a set of actions $\mathcal{A}$ i.e., layer implementations. RL suits well the specifications of the problem that we address in this work. Latency, accuracy or memory represent clear reward function given by the environment that we aim to explore: a Deep Neural Network.  \par

\subsubsection{\textbf{State Space}}
The agent samples sequentially a new set of implementations for the network, layer by layer. The state space is defined as a tuple of the parameters that specify the execution implementation of a layer on a target platform, see Table~\ref{state_space}. All implementations are defined by an algorithm, its configuration  format, a data type, a layout and a HW processor. The agent chooses one implementation from the set of acceleration libraries given the current layer type. Based on the action, the agent moves to another state and the process is repeated until the end of the network.

\subsubsection{\textbf{Exploration Strategy}}
Similar to Baker et al.~\cite{baker2016designing}, we have implemented an $\epsilon$-greedy strategy~\cite{mnih2015human} which trades off between exploitation and exploration. The agent starts mainly exploring the design space (random actions) to sample the diverse possibilities ($\epsilon=1$). We slowly decrease $\epsilon$ over the episodes for the agent to select the best actions and finally learn an optimal path: full exploitation ($\epsilon=0$). In addition, we have added an experience replay~\cite{lin1992self}, a technique that reuses past experiences, to help the action-value function converge faster. After each episode, a batch of past experiences are sampled and presented to the agent. We have set the experience replay's buffer size to 128 following~\cite{baker2016designing}.\par

\subsubsection{\textbf{Reward Function}}
As our main goal is to optimise the deployment of pre-trained DNNs on Cortex-A CPUs, we primarily focus on the latency as our main optimisation goal while setting the accuracy loss and memory consumption as hard constraints. The objective is to maximize the total reward, in this case, minimize the inference time. Although we initially used the total network inference time as unique reward signal, we have added rewards at each step for better convergence (Reward Shaping~\cite{Wiewiora2010}). Hence, each state receives as reward its own layer inference time but reversing the sign, e.g. 0.01ms $\Rightarrow$ -0.01ms. Thanks to the Q-learning update rule, each layer also receives Q-knowledge from the best following state. Therefore, the agent is able to combine both sources of knowledge, look ahead and avoid local minima due to penalties introduced by incompatibility between layers, see Fig.~\ref{layers}.

\section{\textbf{Automated DSE for Deployment of DNNs}} \label{qs-dnn}
We name \textit{QS-DNN (Q-based Search)} to our automated DSE framework for the deployment of DNNs. We envision an industrial deployment scenario where QS-DNN is a one-off process performed only once for a given problem (deep learning task) and target platform. The aim of QS-DNN is to automatically optimise the inference of any DNN on an embedded system before the application is finally deployed on the target setup for real use. Besides, thanks to having an automated DSE tightly coupled with the inference engine, we can also obtain interesting trade-offs in terms of accuracy, memory and latency for specific-problem applications.

\begin{figure}[t!]
   \centering
   \includegraphics[width=0.9\linewidth]{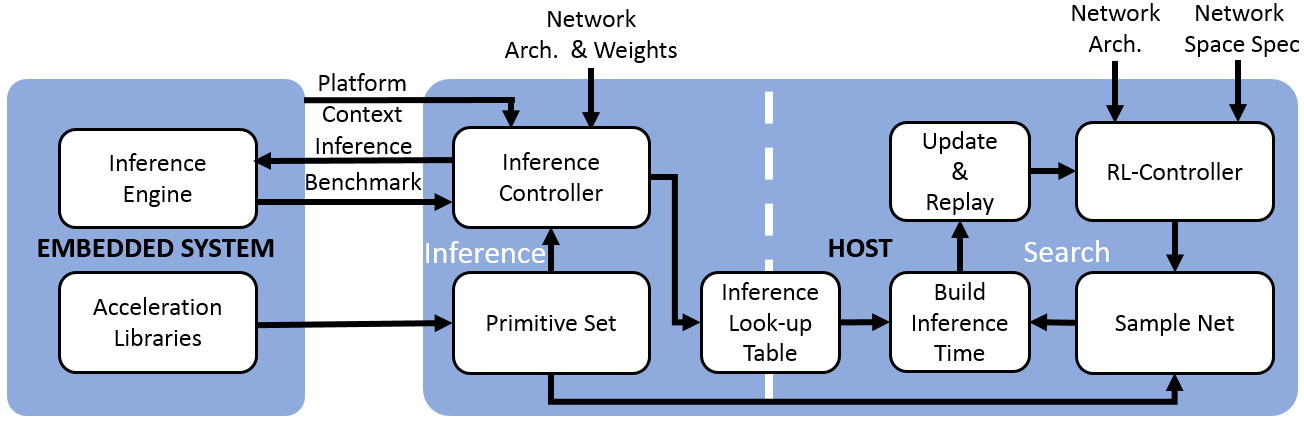}
   \caption{\small\textbf{Architecture of QS-DNN.} Complete flow: Inference on an embedded on the left, RL-based learning on the right.}
   \label{arch}
   \vspace{-0.4cm}
\end{figure}

The process is composed of three phases: \textit{a)} inference of the DNN on the embedded system to obtain empirical measurements: latency and memory, \textit{b)} RL-based search to explore the design space and learn optimised solutions and, \textit{c)} inference of the learnt solutions to obtain accuracy measurements. We have separated the phases (Fig.~\ref{arch}) to avoid inferring on the embedded system each possible solution of the space search, which would significantly slow down the process. Finally, we obtain the Pareto optimal solution based on latency optimisation with accuracy and memory as constraints. \par

\subsection{\textbf{Metrics Collection (1/3)}}
We employ LPDNN, the deep learning inference framework described in Section~\ref{engine}, to obtain real measurements although the search could be also applied to any other framework. We employ LPDNN's acceleration libraries which provide implementations that may leverage the optimisations at network, algorithm and primitive level. The objective of this phase is to measure the costs of all possible edges in the graph, i.e., layer implementations and compatibility conversions, to build a look-up inference table for the search engine. 

Thus, the inference controller goes over each acceleration library and benchmarks each implementation\footnote{Each implementation is inferred for 20 images and the mean is calculated.}, one at a time, in all those layers where the library is able to implement such implementation. Therefore, we only need to infer the whole network on the embedded platform as many times as different global implementations
exist. In each inference, the execution time and memory consumption for each layer are measured. Once all the implementations have been benchmarked, we profile the compatibility conversions for those edges that have not been seen during the first round of benchmarks. Thus, 
we include a second rounds of benchmarks where several implementations are combined for data type and layout transformation as well as for data transfers between different processor (if needed).

\subsection{\textbf{Search Engine (2/3)}}
The search space and the conditions of the search can be defined for each network. They specify the behavior of the agent: number of episodes for each $\epsilon$, learning rate, discounted factor and replay buffer's size. We have set the learning rate to 0.05 and discounted factor to 0.9 to give slightly more importance to short-term rewards. Once the metrics collection phase has finished, the Q-learning -based search begins and proceeds as shown in Algorithm~\ref{alg:QS-DNN}. 

\begin{algorithm}[t!]
  \caption{QS-DNN - Search}\label{alg:QS-DNN}
  \begin{algorithmic}[1]
    \State $\epsilon \gets \epsilon_{new}$
    \While{Learned Episodes $<$ Episodes($\epsilon$)}
      \State Reset Path
      \While{Layer $\neq$ End Layer($\epsilon$)}
        \If{Generate Random $< \epsilon$}
          \State Action $\gets$ Q-values(Random)
        \Else{}
          \State Action $\gets$ Q-values(Max)
       \EndIf
       \State Layer $\gets$ Next Layer
     \EndWhile\label{euclidendwhile}
     \State Check for Incompatibility
     \State Compute Inference Time
     \State Experience Replay \& Update (eq.~\ref{bellman}) 
  \EndWhile\label{layer}
\end{algorithmic}
\end{algorithm}

\begin{figure}[t]
  \vspace{-0.2cm}
   \includegraphics[width=0.9\linewidth]{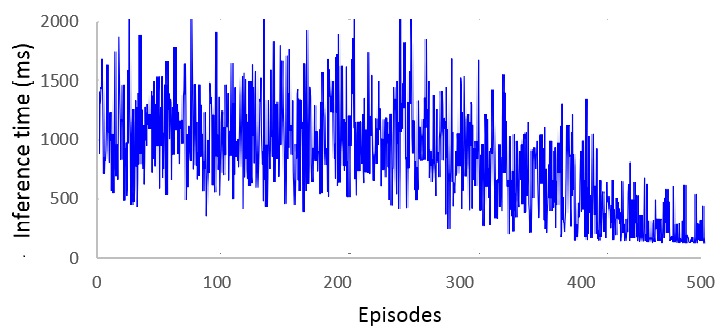}
  \caption{\small\textbf{RL search}. Example of a 500-episode search where the 250 first episodes are fully exploration. From there on, $\epsilon$ is decreased by 0.1 towards exploitation after every 25 episodes.}
  \label{fig:sub1}
  \vspace{-0.5cm}
\end{figure}

First, $\epsilon$ is retrieved from the specifications as well as the number of episodes for such $\epsilon$. In all experiments, \SI{50}{\percent} of the total episodes correspond to full exploration and \SI{5}{\percent} to any other $\epsilon$ from 0.9 to 0.1. By these means, the agent obtains enough knowledge from the environment before starting exploitation, see Fig.~\ref{fig:sub1}. \par

For each episode, the agent samples sequentially a new set of implementations based on the $\epsilon-$strategy. Once the network's configuration is set, the engine automatically looks for incompatibilities between layers. At last, the total network inference time is computed by looking up each implementation in the inference table and summing up the values of all layers. If any incompatibility has been found between two layers, the extra penalty in time is added to the inference time of the latter layer. Finally, the action-value function is updated with the current reward and stored for experience replay. When the number of episodes for a given $\epsilon$ has been met, $\epsilon$ is decreased towards exploitation phase. By the end of the search, the engine gives out the most performing configuration and the learning curve that the agent has followed, see Fig~\ref{fig:sub1}.

\subsection{\textbf{Inference of Learnt Solutions (3/3)}}
A drop in accuracy may be caused by quantised or fast-convolution methods. Thus, we perform the accuracy measurements after the search have been performed and benchmark the learnt solutions against a validation dataset. We are only interested in the most performing networks and hence, only benchmark those solutions that are up to \SI{25}{\percent} slower than the fastest learnt solution. Thus, we can speed up the process and obtain Pareto optimal solutions with a strong focus on latency optimisation having accuracy and memory as thresholds, e.g., accuracy drop $<\SI{1}{\percent}$ or memory reduction $> 2\times$.

\section{\textbf{Experimental Results and Discussion}} \label{results}
In this section, we show the optimisation results for the deployment of state-of-the-art DNNs on a range of Arm Cortex-A CPUs. First, we introduce the set of networks and platforms that we have used to validate our experiments. Next, we show the importance of the different individual software optimisations currently available in LPDNN (see Section~\ref{lpdnn-opt}). Then, we present the Pareto optimal solutions from applying the automated design space exploration, introduced in Section~\ref{qs-dnn}, to optimise the deployment of DNNs on Arm Cortex-A CPUs. Finally, we show a comparison between Reinforcement Learning and several other search algorithms when evaluated on our deployment design space.

\subsection{\textbf{Experimental setup and platforms}}
The set of pre-trained networks that we have selected form part of the Imagenet contest (image classification task) as it represents a challenging dataset where optimisations can have a significant effect on metrics such as latency, accuracy, and memory. We evaluate several representative network topologies for resource-constrained devices that allows us to show the range of optimisation and trade-offs on the target platforms: small networks such as SqueezeNet and MobileNetV3-small, slightly more complex networks like MobileNetV2, and reasonably large networks such as MobileNetV3-large and Resnet50. Although we have focused on an image classification task to demonstrate our design, the experiments could also be applied to any other deep learning task.

The range of Arm Cortex-A (Armv8 64-bit) CPU platforms that we have chosen cover the current spectrum of deployment on mobile devices. We divide the experiment into two parts: \textit{i)} For each of the techniques discussed in Section~\ref{lpdnn-opt}, we present benchmark results on the Raspberry Pi 4 (RPI4; Cortex-A72 at 1.5GHz) as a reference platform to show each optimisation. \textit{ii)} For the automated DSE discussed in Section~\ref{qs-dnn}, we show experiments on the Raspberry Pi 4, Nvidia Jetson Nano, CPU only (Cortex-A57 at 1.43GHz) and, Raspberry Pi 3b+ (RPI3; Cortex-A53 at 1.4GHz) to validate the design and optimisations on “LITTLE” and “big” cores:

\begin{itemize}
    \item \textbf{Cortex-A53} is a low-power, highly efficient, core~\cite{Arm-cortex-a53}. Power efficiency is achieved through running in-order~\cite{anandtech-cortex-a53}, hence this core is highly sensitive to instruction order and operand data availability.
    \item \textbf{Cortex-A57} is a higher-performance, and consequently higher-power, out-of-order core~\cite{Arm-cortex-a57}.
    \item \textbf{Cortex-A72} is an update to the Cortex-A57 with improved floating-point and memory performance~\cite{anandtech-cortex-a72}
\end{itemize}

All inferences are performed identically: using a single-thread, calculating the average of twenty inferences after an initial (discarded) warm-up run. The boards were fitted with heatsinks, and the CPU frequency governors were overridden to achieve the maximum possible performance. To ensure that the platform does not overheat, triggering thermal throttling, we monitored the platforms and sampled the hardware monitors once per second. We execute each DNN implementation without external loads, i.e., resources are not shared among several processes, emulating a mobile device scenario, e.g., an application including AI capabilities executes the DNN execution, the most computational-intensive task, exclusively in one of the cores while the other cores can perform general-purpose tasks. Nonetheless, the scenario where multiple DNNs run concurrently is a very interesting topic where either multiple (RL) algorithms could run concurrently to find the best common deployment or one single one orchestrates the deployment of all DNN. We leave this problem as future work as it could be the subject of a paper itself.

\begin{figure}[t]
   \includegraphics[width=1\linewidth]{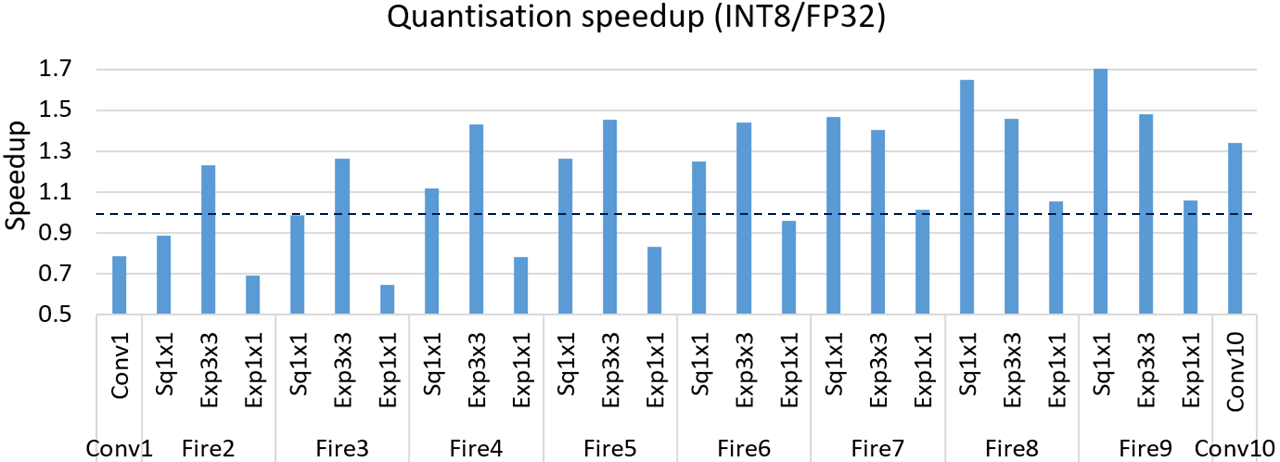}
  \caption{\small\textbf{Quantisation optimisation.} Speedup of INT8 over FP32 for SqueezeNet-V1's convolutions (the higher, the better).}
  \label{fig:quantisation}
  \vspace{-0.2cm}
\end{figure}

\subsection{\textbf{Optimisation results}} \label{opt-results}
Next, we demonstrate the improvement in performance due to the software optimisations explained in Section~\ref{lpdnn-opt}.

\subsubsection{\textbf{Network Optimisation}}
We show three different optimisations at this level: \par
\textbf{Quantisation:} We compare the performance of DNN layers when they are deployed employing INT8 arithmetic instead of the baseline FP32 operations. Although INT8 variables can be packed into 32-bit operations and, therefore, achieve a theoretical $4\times$ arithmetic-intensity improvement, the real uplift in performance may not be as much. For instance, Fig.~\ref{fig:quantisation} shows INT8 performance uplift for SqueezeNet's convolutions, where the highest layer improvement goes no higher than $1.7\times$ and the overall network speedup (including all layers) is $1.24\times$. These results are roughly in line with~\cite{jacob2018quantization}, where speedups of $1.13\times$ and $1.2\times$ are obtained by using INT8 on small and big cores of the Snapdragon~835 respectively.

The relatively low improvement with respect to FP32 is due to several factors: \textit{i)} the need to perform several instructions to multiply and accumulate four 8-bit integer terms into a single 32-bit integer result\footnote{Newer versions of the Arm ISA include the \texttt{DOT} and \texttt{MMLA} instructions to mitigate this~\cite{arm-dot-instructions,arm-mmla-instructions}.}, \textit{ii)} the need to requantise the activations back to INT8 (scale and offset) after each layer and, \textit{iii)} the existence of very performant FP32 primitives thanks to an intense optimisation over the years. \par

\begin{figure}[!tbp]
  \centering
  \subfloat[\textbf{Static fusion:} Bnorm and Scale layers into previous convolutions for MobileNet-V2.]{\includegraphics[width=0.48\linewidth]{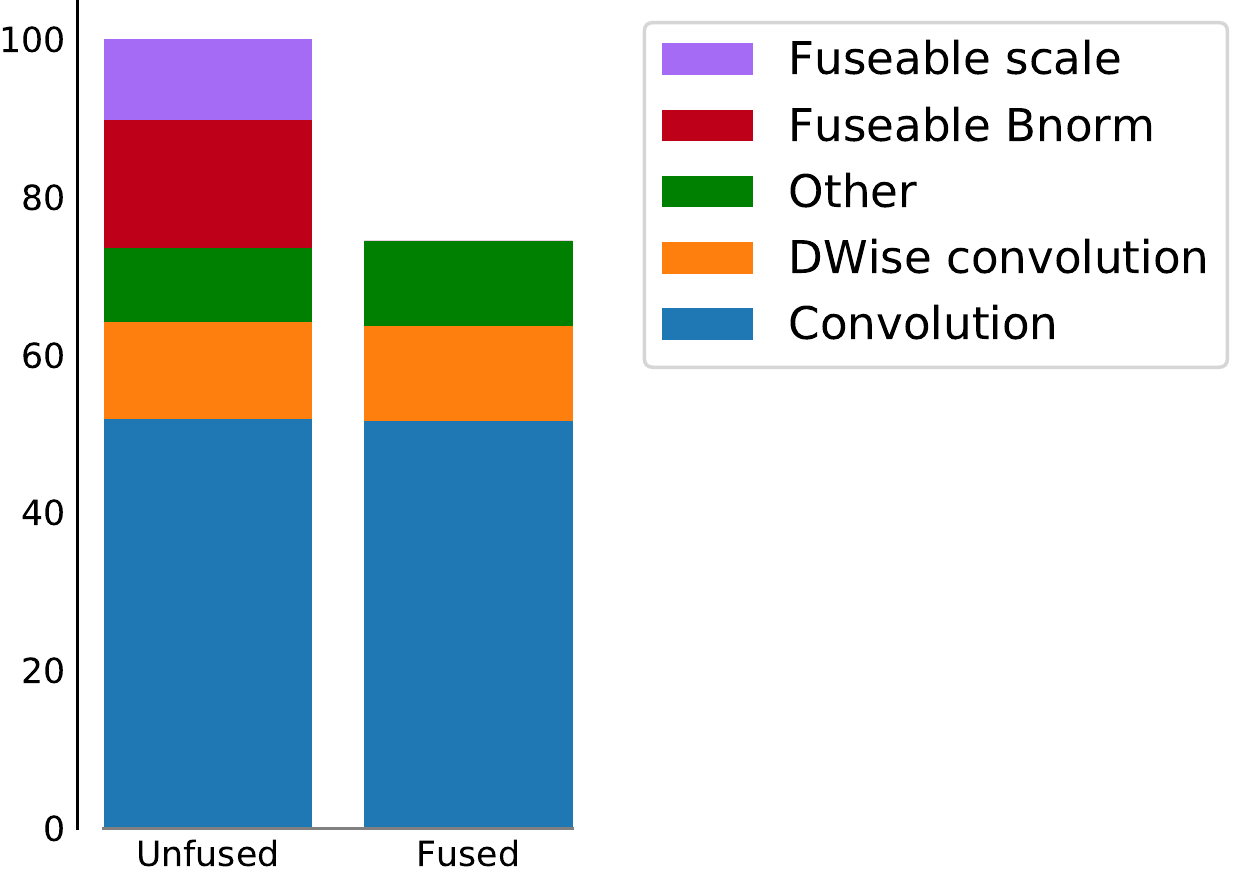}\label{fig:static_fusion}}
  \hspace{0.1cm}
  \subfloat[\textbf{Runtime fusion:} Activation and Concat layers into previous convolutions for SqueezeNet-V1.]{\includegraphics[width=0.48\linewidth]{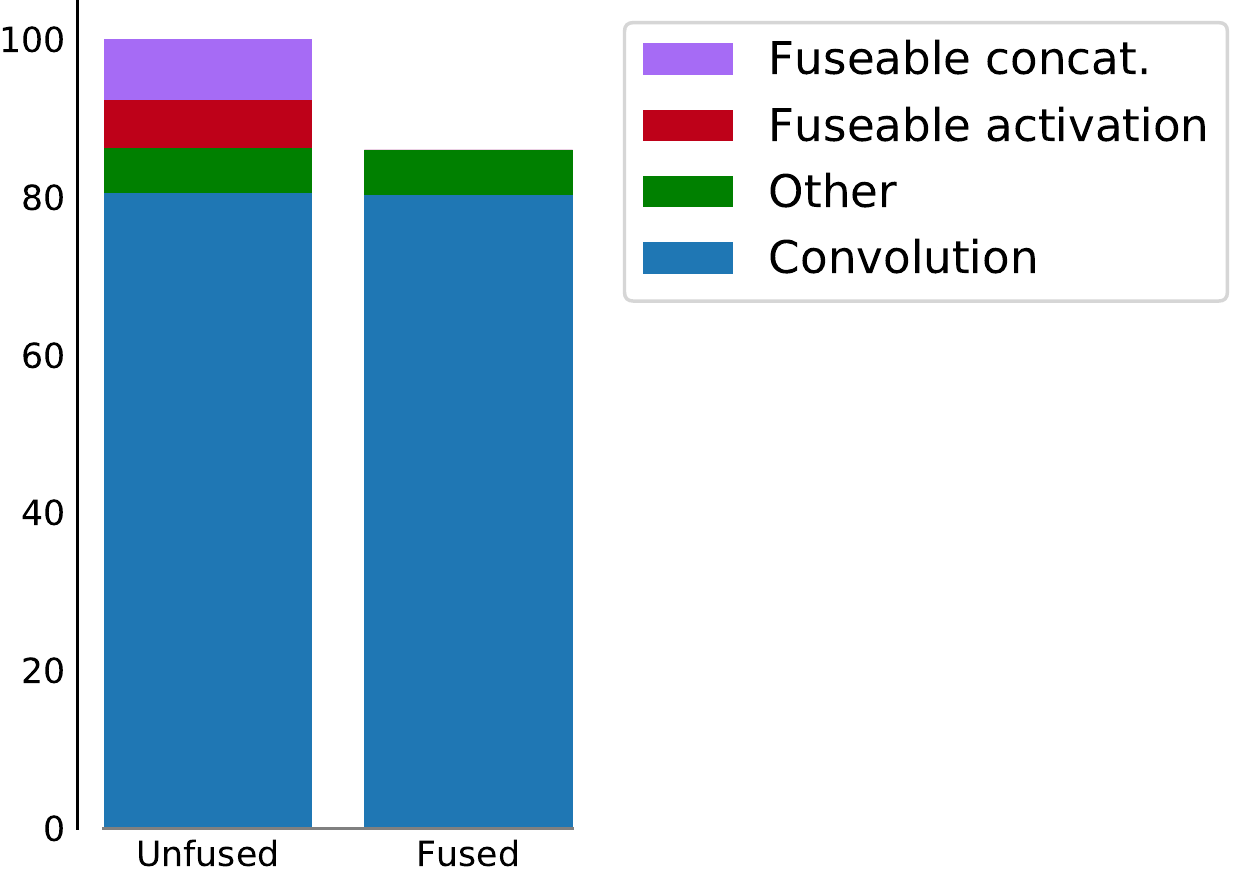}\label{fig:dynamic_fusion}}
  \caption{\small\textbf{Layer fusion optimisation.} Portion of overall execution time. Normalised to non-fused performance (the lower, the better).}
  \label{fig:fusion}
  \vspace{-0.4cm}
\end{figure}

\textbf{Layer fusion:} Fig.~\ref{fig:fusion} shows the performance improvement obtained by static and runtime layer fusions. Fig.~\ref{fig:static_fusion} depicts the static fusion of Bnorm and scale layers into previous convolutions for MobileNet-V2, obtaining a \SI{25}{\percent} improvement in latency. Such significant improvement is mostly due to the lack of optimisation on both fused layers as this graph reduction is a common approach in many inference frameworks. Fig.~\ref{fig:dynamic_fusion} exhibits the fusion of activation and concatenation layers into preceding convolutions for SqueezeNet, achieving an improvement of \SI{15}{\percent} in latency due to the reduction of memory accesses (for only a modest increase in the time spent performing convolution ($<\SI{1}{\percent}$). \par

\textbf{Memory optimisation:} Table.~\ref{memory-opt-results} displays the total dynamic memory allocated over the execution of various DNNs in terms of weights, activations, and other (structures, buffers, or code). We can observe that, while the activations account for a small portion in Resnet50, they consume most of the memory allocated in Squezeenet and MobileNet-V2. Thus, we show the reduction in memory achieved in the allocation of the activations by in-place and memory pool techniques. The use of in-place technique achieves a noticeable reduction, especially on the ReLu layers, which varies from \SI{16.1}{\percent} (Resnet50) to \SI{37.2}{\percent} (MobileNet-V2). Further, memory pool technique accomplishes a remarkable memory reduction that goes from \SI{80}{\percent} (SqueezeNet) to \SI{88.1}{\percent} (MobileNet-V2) thanks to the reuse of memory across layers. Further memory optimisations can be achieved by applying quantisation which would provoke a $4\times$ reduction in the memory allocated for the weights and the remaining activations. \par

\begin{table}[t!]
\begin{center}
\begin{tabular}{lSSS}
& {SqueezeNet-V1} & {MobileNet-V2} & {Resnet50}  \\
& {(\SI{48}{\mebi\byte})} & {(\SI{96}{\mebi\byte})} & {(\SI{425}{\mebi\byte})}  \\
\toprule
Weights / \si{\percent} & 9.7 & 14.1 & 23.1 \\
Activations / \si{\percent} & 60.6 & 70.2 & 23.1 \\
Other / \si{\percent} & 29.7 & 15.7 & 53.8 \\
\midrule
Act. (Original) / \si{\percent} & 100.0 & 100.0 & 100.0 \\
Act. (In-place) / \si{\percent} & 70.4 & 62.2 & 83.9 \\
Act. (Mem pool) / \si{\percent} & 20.0 & 11.9 & 13.0 \\
\bottomrule
\end{tabular}
\end{center}
\caption{\textbf{Memory optimisation.} Upper part: total memory allocated for the weights, activations and other (code, struct, etc.) Lower part: memory allocated for the activations (normalised to activation (original)).}
\label{memory-opt-results}
\end{table}

\subsubsection{\textbf{Algorithm Optimisation}}
\begin{figure}[t]
   \includegraphics[width=1\linewidth]{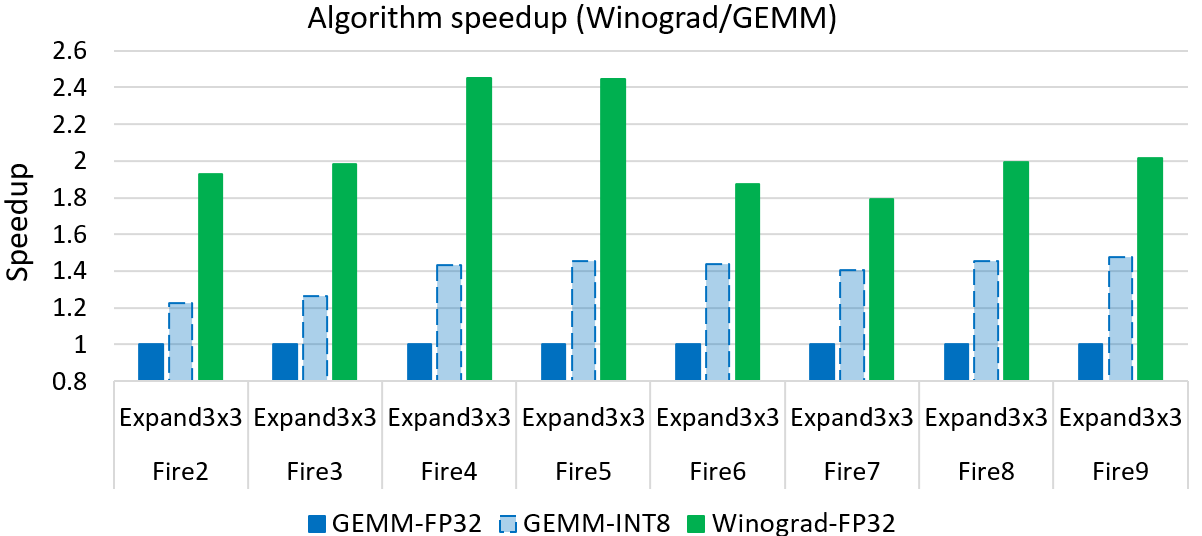}
  \caption{\small\textbf{Algorithm optimisation.} Speedup of Winograd over GEMM-FP32 for SqueezeNet-V1 (the higher, the better).}
  \label{fig:algorithm-results}
  \vspace{-0.4cm}
\end{figure}

We present the comparison between GEMM and Winograd algorithms - employing LPDNN-Arm library - as these two represent the most common algorithms for convolutions on CPU deployment. Fig.~\ref{fig:algorithm-results} illustrates the speedup achieved by Winograd-FP32 over GEMM-FP32 for SqueezeNet's 3x3 convolutions (Winograd not available for k=1x1 in LPDNN). We can observe that Winograd clearly outperforms GEMM-FP32, accomplishing an uplift of up to $2.5\times$. Likewise, Winograd achieves a speedup of up to $3.9\times$ for Resnet50, especially in the first convolutions, which are the most computing-intensive (Fig.~\ref{fig:algorithm-results-resnet} in Appendix). 

Further, we have included GEMM-INT8 performance in both figures to demonstrate that early design decisions may lead to sub-optimal solutions, e.g., selecting the use of quantisation at network level without considering the underneath algorithm level. In this case, selecting uniform quantisation, e.g., INT8 data type for the whole network, discards Winograd algorithm as, to date, it is only available in FP32. Thus, a homogeneous INT8 network would underperform against a mixed-precision network, including Winograd.

\subsubsection{\textbf{Primitive Optimisation}}
\begin{figure}[t]
   \includegraphics[width=1\linewidth]{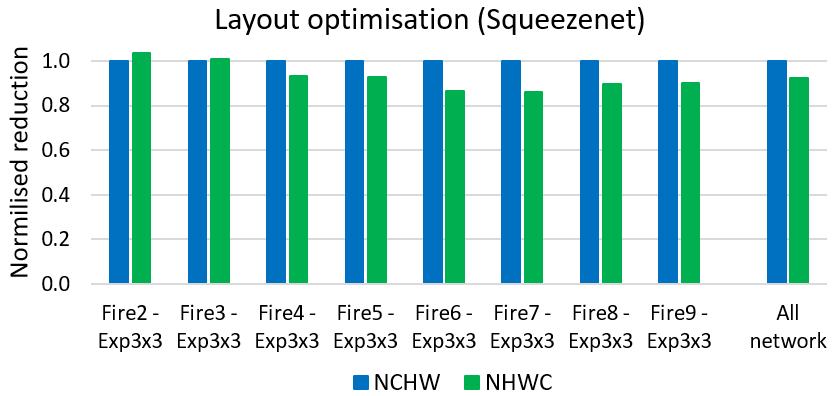}
  \caption{\small\textbf{Primitive optimisation.} Layout performance comparison between NCHW and NHWC for SqueezeNet's most computational-intensive layers. (the lower, the better).}
  \label{fig:primitive-results-squeeze}
\end{figure}

We show the influence of data layout for vectorised methods employing the GEMM algotithm. Fig.~\ref{fig:primitive-results-squeeze} presents the most computational-intensive convolutions of SqueezeNet (kernel=3x3). We can observe that NCHW performs slightly better in the first layers, probably due to a lower number of channels in this stage of the network and, therefore, lower data reuse under the NWHC layout. Yet, NHWC broadly outperforms NCHW on 3x3 kernels, and also on 1x1 kernels, which, having a lower degree of data reuse across the plane of the convolution, are notably more performing under the NHWC layout. Overall, NHWC achieves a reduction in the latency of \SI{8}{\percent} throughout the network.

Another substantial improvement is the optimisation of the first convolution, which typically accounts for one of the most time-consuming layers. Input data is generally organised in NCHW-order while, by contrast, many performant convolution primitives prefer an NHWC layout (as we saw previously). However, converting from NCHW to NHWC (particularly where only three channels of data are concerned) to match the preferred input layout of the first convolution is relatively costly and hence undesirable. We can avoid this by taking input data in NCHW and producing NHWC directly -- avoiding additional rearrangement. This is achieved simply by providing an \texttt{im2row} routine specialised for NCHW-order data and a low channel count.

\begin{figure}[t]
   \includegraphics[width=1\linewidth]{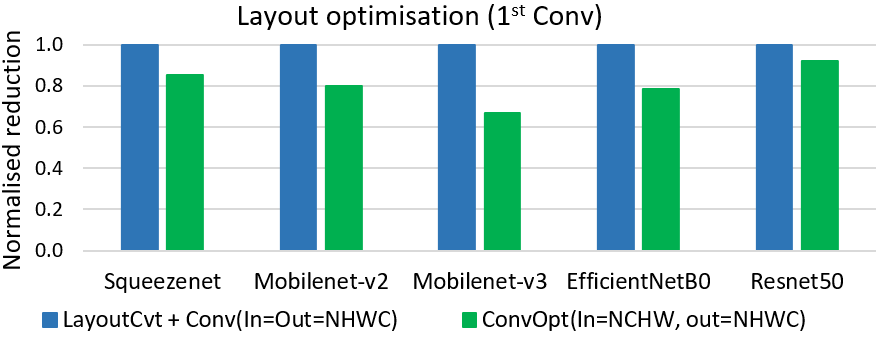}
  \caption{\small\textbf{Primitive optimisation.} Layout optimisation for the $1^{st}$ convolution by avoiding layout rearrangements thanks to an \texttt{im2row} specialised routine. (the lower, the better).}
  \label{fig:primitive-results-all}
  \vspace{-0.4cm}
\end{figure}

\subsection{\textbf{Automated Design Space Exploration (DSE)}} \label{dse-results}
We aim to show the trade-offs between latency, accuracy, and memory footprint when all the different levels of software optimisations shown in Section~\ref{opt-results} are available and may be applied, e.g., employing either quantised, or fast-convolution methods may have a substantial impact in all three metrics. We illustrate the DSE and the optimisations achieved by QS-DNN by providing two Pareto fronts: latency-accuracy and latency-memory, where we present the achievable performance based on different degrees of accuracy and memory. 

To show the optimization improvements, we display the following \textit{interesting} points on the graphs:
\begin{itemize}
    \item \textbf{Ref-FP32}: We take LPDNN coupled with Openblas library as reference implementation since it is a well-known and standard library for industrial deployment. Ref-FP32 employs GEMM-based methods under NCWH layout for convolutions and fully connected layers and direct methods for any other layer.
    \item \textbf{Opt-FP32}: Solution found by QS-DNN when only FP32 implementations are allowed, i.e., no quantised methods.
    \item \textbf{INT8}: Fully INT8 deployment implementation.
    \item \textbf{Pareto front}: Set of points that are not dominated by any other implementation based on the two given objectives.
\end{itemize}

For the sake of fairness, \textit{static layer fusion} (graph reduction) and \textit{memory pool} optimisations are always ON for all implementations. 

\subsubsection{\textbf{Latency-Accuracy}}

\begin{figure}[t]
   \includegraphics[width=1\linewidth]{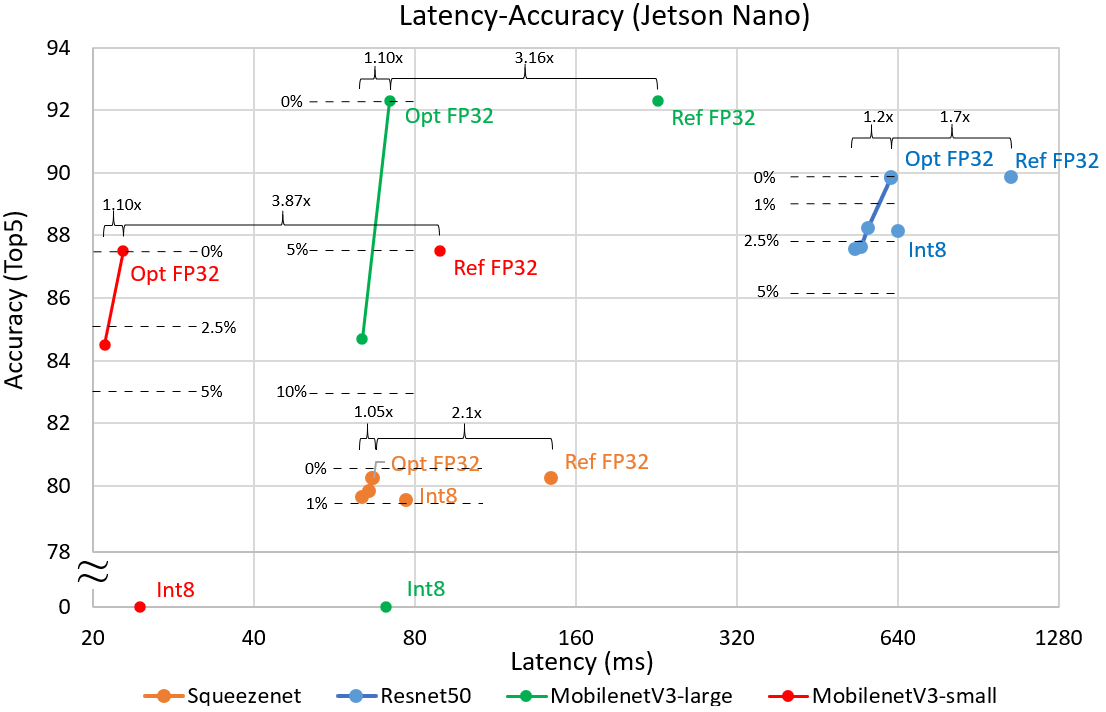}
  \caption{\small\textbf{Latency vs Accuracy trade-off.} Automated DSE on Jetson Nano (A-57) showing the most \textit{interesting} points. Solid line represents the Pareto front. X-axis is log2 based.}
  \label{fig:accuracy-nano}
\end{figure}

\begin{figure}[t]
   \includegraphics[width=1\linewidth]{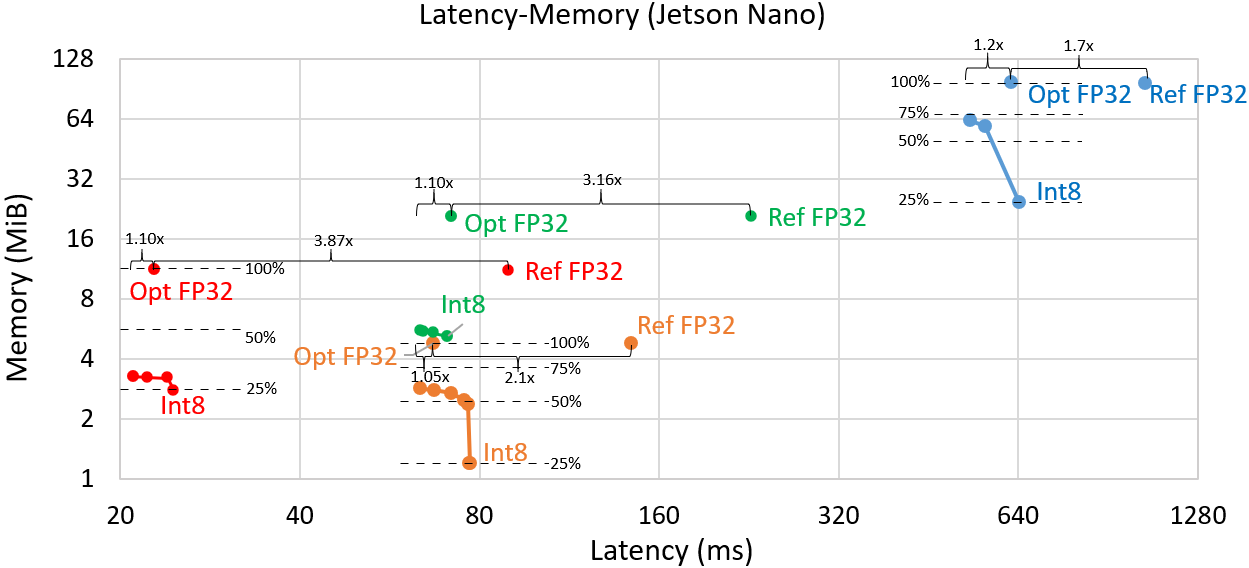}
  \caption{\small\textbf{Latency-Memory trade-off.} Automated DSE on Jetson Nano (A-57) showing the most \textit{interesting} points. Solid line represents the Pareto front. X- and Y-axes are log2 based.}
  \label{fig:memory-nano}
  \vspace{-0.4cm}
\end{figure}

\begin{figure}[t]
   \includegraphics[width=1\linewidth]{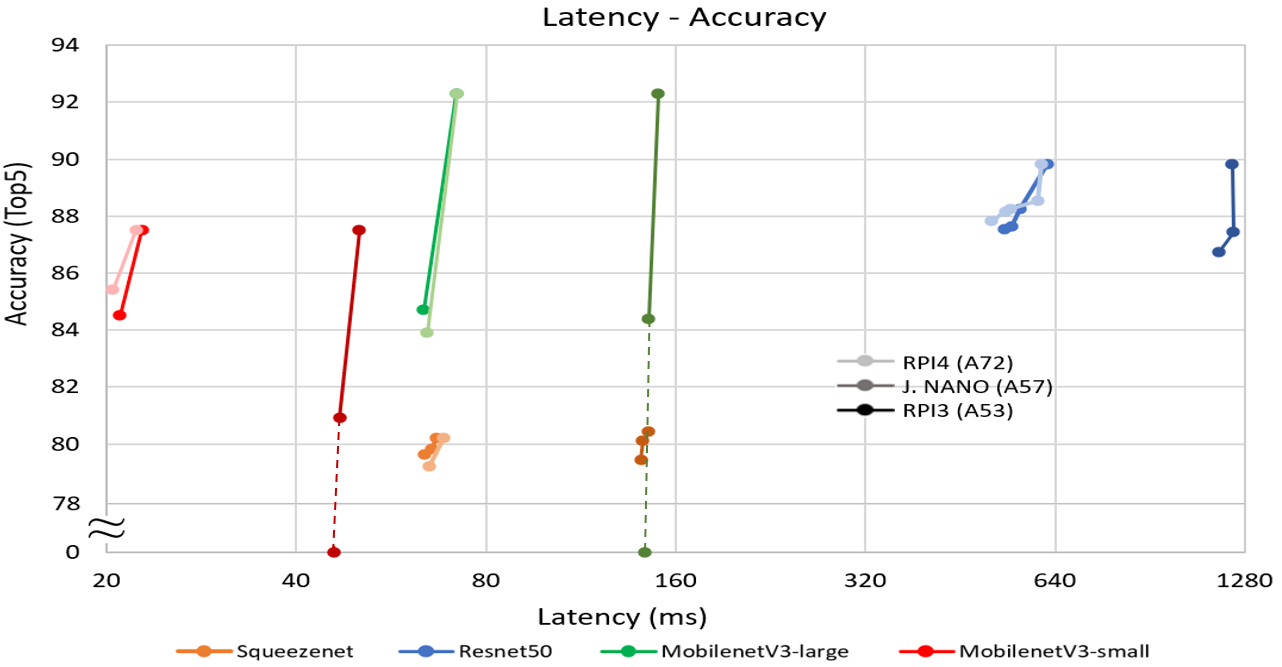}
  \caption{\small\textbf{Latency-Accuracy trade-off.} Pareto front comparison between the RPI3, RPI4, and Jetson Nano. X-axis is log2 based.}
  \label{fig:accuracy-all}
\end{figure}

\begin{figure}[t]
   \includegraphics[width=1\linewidth]{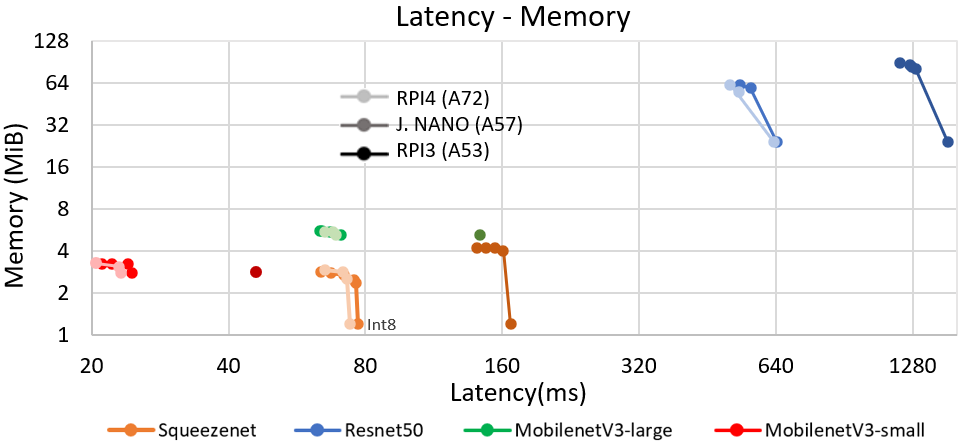}
  \caption{\small\textbf{Latency-Memory trade-off.} Pareto front comparison between the RPI3, RPI4, and Jetson Nano. X- and Y-axes are log2.}
  \label{fig:memory-all}
\end{figure}

\begin{table*}[t!]
\begin{center}
\begin{tabular}{lcccccccc}
 & {Ref-FP32} & {RL-A} & {RL-L} & {Dj} & {A*} & {DS+} & {DS} & {RS}  \\
\toprule
Complexity (worst case) & & $\mathnormal{O(le)}$ & $\mathnormal{O(le)}$  & $\mathcal{O(V\log{}V+E)}$  & $\mathcal{O(V\log{}V+E)}$  & $\mathcal{O(V)}$  & $\mathcal{O(V)}$  & $\mathnormal{O(le)}$ \\
\midrule
Resnet50's latency (ms) & 1045 & 623.0 & 532.2 & 532.2 & 533.3 & 533.7 &  580 & 900 \\
\# Considered states (K) & / & 56.8 & 56.8 & 4.1 & 3.4 & 0.91 &  0.91 & 56.8 \\
Accuracy (\%) & 89.8  & 89.8 & 87.6 &  87.6 & 87.6 & 87.6 & 87.4 & 88.8  \\
\midrule
MobilenetV3-L's latency (ms) & 228.4 & 70.3 & 63.7 & 63.7 & 63.8 & 64.8 & 78.4 & 153.7 \\
\# Considered states (K) & / & 65.14 & 65.14 & 3.81 & 3.62 & 0.92 & 0.92 & 65.14 \\
Accuracy (\%) & 92.3 & 92.3 & 84.7 & 84.7  & 84.5 & 84.7 & 89.1 & 91.0  \\
\bottomrule
\end{tabular}
\end{center}
\caption{\small\textbf{Search algorithms' evaluation (latency optimisation).} The solution found for the deployment of Resnet50's and MobilenetV3-Large on the Jetson Nano is given in milliseconds together with the number of states considered during the search and the final accuracy.}
\label{search-results}
\vspace{-0.4cm}
\end{table*}

Fig.~\ref{fig:accuracy-nano} illustrates the most \textit{interesting} points found by QS-DNN while performing a DSE for the range of networks on the Jetson Nano. Dotted lines represent the loss in accuracy (\SIlist{1;2.25;5;10}{\percent}) with respect to the \textit{Ref-FP32} implementation while braces expose the performance gains.
\par
Generally, we can observe that the DSE provided by QS-DNN for \textit{Opt-FP32} clearly outperforms \textit{Ref-FP32} from $1.7\times$ (Resnet50) to $3.87\times$ (MobileNetV3-small) with no loss in accuracy. We can explain this significant improvement mainly due to the selection of Winograd convolutions for SqueezeNet and Resnet50 and, optimised depthwise convolutions for MobileNets. Besides, techniques such as runtime layer fusion and the optimisation of layouts through the network make a meaningful impact in performance. 

By allowing QS-DNN to select quantised methods, further improvement can be achieved with only a small drop in accuracy: up to $1.05\times$ and $1.2\times$ increase in performance with under \SI{1}{\percent} and \SI{3}{\percent} drop in accuracy for SqueezeNet and Resnet50, respectively. MobileNets, on the other hand, are significantly more sensitive to quantisation and their accuracy drops drastically when quantised methods are employed. This sensitivity to quantisation is largely due to depthwise convolutions, as each channel represent an independent kernel and may have very different ranges. LPDNN's acceleration libraries only support layer quantisation, i.e., one range or scale per layer, and selecting quantised methods for depthwise layers clearly hurts the accuracy (see the fully \textit{INT8} solution on Fig.~\ref{fig:accuracy-nano}). However, thanks to the DSE, we can find mixed-precision solutions that for instance, bring up to $1.10\times$ 
increase in performance with a modest drop in accuracy of around \SI{3}{\percent} 
for MobileNetV3-small.

If we compare the Pareto fronts of the networks on the different platforms, we can see in Fig.~\ref{fig:accuracy-all} that there is not a significant difference in latency or accuracy between the RPI4 and the Jetson Nano while the RPI3, containing a ``LITTLE'' Arm core, performs around $2\times$ slower. Interestingly, no fully \textit{INT8} solution forms part of the Pareto front on the RPI4 and Jetson Nano platforms. This indicates that fully quantised networks may not always be optimal in terms of latency  due to a lack of support for certain primitives and the need for requantisation after each layer.

\subsubsection{\textbf{Latency-Memory}}
As we described earlier, the \textit{memory pool} optimisation for the activations is always ON and achieves over \SI{80}{\percent} reduction in memory allocation. Hence, we can only achieve further reductions of memory by reducing the size of the weights, e.g., through quantisation.
Fig.~\ref{fig:memory-nano} shows the most \textit{interesting} points found by QS-DNN while performing a DSE for the range of networks on the Jetson Nano. Dotted lines represent the memory consumption for the weights (\SI{75}{\percent}, \SI{50}{\percent}, and \SI{25}{\percent}) with respect to the \textit{Ref-FP32} implementation while braces expose the performance gains. 

Building from the Latency-Accuracy graph, we see that the Latency-Memory Pareto fronts are convex rather than linear. While FP32 layers are preferred in the Latency-Accuracy DSE  due to their higher precision, in this case, quantised methods are favoured due to their lower memory footprint, being the fully \textit{INT8} solution the lowest point of the Pareto front (\SI{25}{\percent}) for all networks. From the lowest point, the Pareto rises and turns towards less latency as some more performing FP32 algorithms are picked, e.g., Winograd, which increases the memory up to the \SI{58}{\percent} and \SI{65}{\percent} ratio for SqueezeNet and Resnet50, respectively.

Distinctly, MobileNets' Pareto fronts rise less and tend to remain close to \SI{25}{\percent} as they do not contain any Winograd implementation and quantised methods are more efficient. Overall, if we combined the optimisations in memory from both the weights and activation, the total memory reduction can go up to $1.9\times$ and $2.3\times$ for small networks like SqueezeNet and MobileNetV3-small and, up to $1.6\times$ and $2.5\times$ for larger networks such as Resnet50 and MobileNetV3-large. 

From the comparison between the Pareto fronts on the different platforms (Fig.~\ref{fig:memory-all}), we can draw that the RPI4 performs slightly better than the Jetson Nano. This may indicate, given the similar FP32 performance, that quantised methods are more performing on the A72 core than the A57 counterpart. Interestingly, the Pareto front of MobileNets on the RPI3 consists of one point, the \textit{INT8} solution. This can be explained by looking at the Latency-Accuracy Pareto front of MobileNets where \textit{INT8} solutions are the fastest ones (despite the poor accuracy), which denotes the efficiency of quantised methods for this network topology on the A53 core.

\subsubsection{\textbf{Discussion}}
Thanks to the automatic DSE provided by QS-DNN, we can quickly analyse several DNNs on a range of embedded platforms and find a suitable solution for a given problem. Thus, analysing the previous experiments, we can observe that the recent MobileNetV3s are far more efficient than SqueezeNet and Resnet50, e.g., MobileNetV3-small performs $3\times$ faster than SqueezeNet (on the Arm Cortex-A57) and is and over \SI{8}{\percent} more accurate while having the same memory footprint. Likewise, we could select the most suitable platform for a given problem based on latency, memory or energy constraint.

\subsection{\textbf{Search Algorithms' Evaluation}}
Finally, we compare Reinforcement Learning with several well-known search algorithms when evaluated on our deployment design space as described in Section \ref{sec:learning-based-search-engine}:
\begin{itemize}
    \item \textbf{Random (RS)}: Random selection of sequential implementations throughout the network - no learning or optimisation. The search starts over from the beginning each time and is repeated for a given number of episodes. 
    \item \textbf{Direct (DS)}: One-go greedy search that evaluates next layer’s implementations and picks the one with the smallest cost (DS). If the transformation penalty is also included in the cost, we name the search DS+. The difference between DS and DS+ will allow us to observe the effect of incompatibilities between layers.
    \item \textbf{Dijkstra (Dj)}: Graph transversal search that finds the shortest path between a node and every other (greedy) \cite{dijkstra1959note}. All neighbors are evaluated at each step and the course of the search is updated when a better path is found. Each node can be only visited once. 
    \item \textbf{A*}: It is a generalisation of Dijkstra's algorithm that decreases the searching space by adding heuristics to the cost, helping the search converge faster \cite{hart1968formal}. We define the heuristic for each neighbour as the minimum cost from all implementations when looking at the neighbour's neighbours, i.e., two steps ahead from the current node. 
\end{itemize}
Table \ref{search-results} presents the searching algorithms' results when evaluated on Resnet50's and MobilenetV3-Large's deployment design spaces (latency optimisation). 
We take Ref-FP32 as the reference, as described in Section \ref{dse-results}, and show two points from RL's Pareto front, trading-off latency and accuracy: \textit{1) RL-L} which emphasises on latency and \textit{2) RL-A} which emphasises on accuracy.

Taking only the latency into account, we can observe in Resnet50's deployment that RL, having a time complexity of $\mathnormal{O(le)}$ where $\mathnormal{l}$ is the number of layers and $\mathnormal{e}$ is the number of episodes in the search, accomplishes an uplift of $2\times$ over Ref-FP32. Dijkstra, having a worst-case complexity of $\mathcal{O(V\log{}V+E)}$, where $\mathcal{V}$ is the number of vertices and $\mathcal{E}$ is the number of connections, finds the same solution as RL while reducing the number of visited states by over $13\times$. A* finds almost the same solution and decreases the searching cost by $1.2\times$ over Dijkstra, thanks to the heuristic's extra information.

DS+, having a complexity of $\mathcal{O(V)}$, obtains a very slightly worse solution while dropping the searching cost by $3.7\times$ compared to A*. If we compare DS+ and DS, we can conclude that the transformation penalties between incompatible implementations have a noticeable effect. Taking them into the search provides an improvement of 8\% and 17\% in time for Resnet50 and MobilenetV3-L, respectively. Overall, Dijkstra, A*, and DS+ find an optimised solution comparable to that found by RL while significantly reducing the searching cost. Therefore, they are preferable and can be used within QS-DNN for single-objective optimisation.

However, these algorithms are not always easily ported to constraint searches, e.g., accuracy loss, as the network's accuracy will only be known after the search's solution is available. Besides, Dijkstra, A*, and DS+ only provide a one-way search, i.e., only one solution is provided at the end. Although Dijkstra's, A*'s, and DS+'s solutions only incur in an accuracy loss of just over 2\% compared to the Ref-FP32 for Resnet50, they suffer a large degradation in accuracy (almost 8\%) for MobilenetV3-L.

By contrast, RL provides an iterative search, yielding as many solutions as given episodes. RL explores the design space and provides a Pareto front solution to trade-off accuracy and latency. Thus, other than RL-L solution for MobilenetV3-L - similar to Dijkstra's - we can also find other design points like RL-A's, with no loss in accuracy and over $3\times$ faster than Ref-F32. Hence, we find RL's search more suitable for our industrial deployment scenario, i.e., a one-off optimisation process (non-critical time) before the final deployment, especially for cases like MobileNet, where Pareto solutions are essential to avoid low-accuracy implementations.

\section{Conclusion and Future Work}
We have analysed the methods to improve the deployment of DNNs across the different levels of software optimisation and introduced the range of techniques provided by LPDNN to optimise DNN inference. Building on this knowledge, we have shown that single optimisation methods may lead to sub-optimal deployment for end-to-end solutions in terms of latency, accuracy, and memory footprint. 

Therefore, we have introduced an automated exploration framework that relies on a Reinforcement Learning search which, combined with LPDNN, enables the deployment of DNN implementations to obtain empirical measurements. Thus, we are able to learn an optimised solution for a given task by automatically exploring the deployment design options on the target embedded device. To validate the design, we have presented a set of results for state-of-the-art DNNs on a range of Arm Cortex-A CPU platforms achieving up to $4\times$ improvement in performance and over to $2\times$ reduction in memory with negligible loss in accuracy with respect to the BLAS floating-point implementation. 

We aim to extend this work to micro-controllers where the resources are very limited, and careful design needs to be performed. Further, we envision extending this work for runtime adaptation of the AI solution by having an online search to improve continuously the latency and memory consumption based on the environment state.

\section{Acknowledgement}
This project has received funding from the European Union's Horizon 2020 research and innovation programme under grant agreement No. 732204 (Bonseyes). This work is supported by the Swiss State Secretariat for Education, Research and Innovation (SERI) under contract number 16.0159. The opinions expressed and argument employed herein do not necessarily reflect the official views of these funding bodies.

\bibliographystyle{IEEEtran}
\bibliography{references}

\appendix

\section{Quantisation}
\begin{figure}[H]
   \includegraphics[width=1\linewidth]{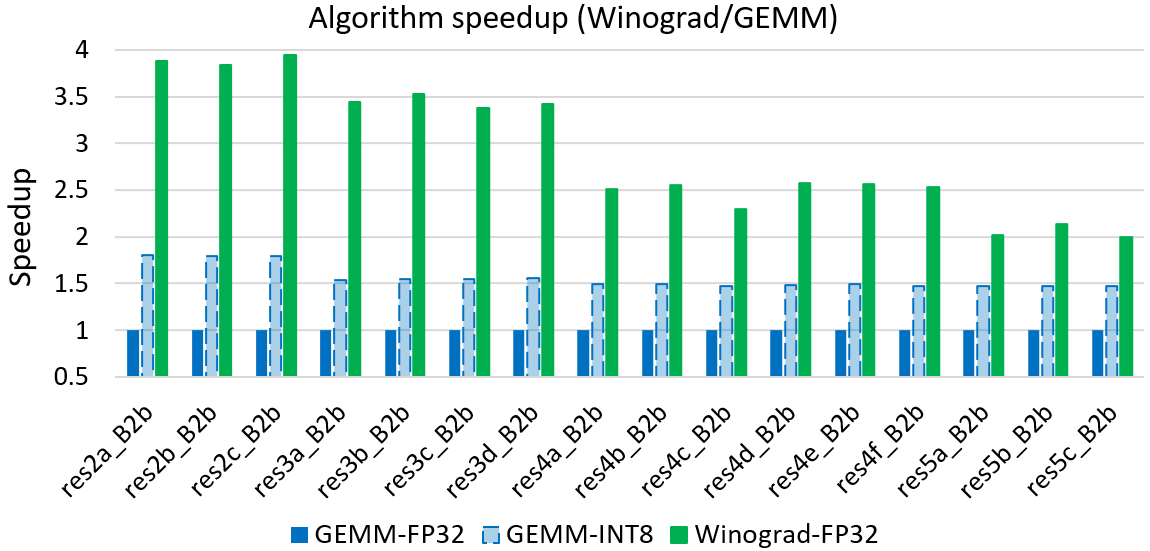}
  \caption{\textbf{Algorithm optimisation.} Speedup of Winograd over GEMM-FP32 for Resnet50' (the higher, the better).}
  \label{fig:algorithm-results-resnet}
\end{figure}

\end{document}